\documentclass{article}

\usepackage{PRIMEarxiv}

\usepackage[utf8]{inputenc} 
\usepackage[T1]{fontenc}    
\usepackage[hyperfootnotes=false]{hyperref}
\usepackage{url}            
\usepackage{booktabs}       
\usepackage{amsfonts}       
\usepackage{nicefrac}       
\usepackage{microtype}      
\usepackage{lipsum}
\usepackage{fancyhdr}       
\usepackage{graphicx}       
\graphicspath{{media/}}     

\usepackage{amsmath}
\usepackage{amssymb}
\usepackage{algpseudocode}
\usepackage{algorithm2e}
\usepackage{empheq}
\newcommand{\D}{\mathbf{D}}
\newcommand{\Q}{\mathbf{Q}}
\newcommand{\Ma}{\mathbf{M}}

\usepackage[toc,page]{appendix}
\usepackage{color,array}
\usepackage{cite}
\usepackage{float}
\title{Non-Intrusive Reduced Models based on Operator Inference for Chaotic Systems
}

\author{
  J.L.S de Almeida, \, A.C. Nogueira Jr., \\
  IBM Research\\
  Hortolândia, SP - Brazil\\
  \texttt{joao.lucas.sousa.almeida@ibm.com, albercn@br.ibm.com}\\
   \And
  A.C. Pires, \,K.F.V. Cid \\
  Department of Mechanical Engineering - State University of Campinas \\
  Campinas, SP - Brazil\\
  \texttt{arthur-pires@live.com, klaus.cid@gmail.com} \\
}

\begin{document}
\maketitle
\let\thefootnote\relax\footnotetext{Copyright \copyright 2022 IEEE-TAI}

\begin{abstract}
This work explores the physics-driven machine learning technique Operator Inference (OpInf) for predicting the state of chaotic dynamical systems. OpInf provides a non-intrusive approach to infer approximations of polynomial operators in reduced space without having access to the full order operators appearing in discretized models. Datasets for the physics systems are generated using conventional numerical solvers and then projected to a low-dimensional space via Principal Component Analysis (PCA). In latent space, a least-squares problem is set to fit a quadratic polynomial operator, which is subsequently employed in a time-integration scheme in order to produce extrapolations in the same space. Once solved, the inverse PCA operation is applied to reconstruct the extrapolations in the original space. The quality of the OpInf predictions is assessed via the Normalized Root Mean Squared Error (NRMSE) metric from which the Valid Prediction Time (VPT) is computed. Numerical experiments considering the chaotic systems Lorenz 96 and the Kuramoto-Sivashinsky equation show promising forecasting capabilities of the OpInf reduced order models with VPT ranges that outperform state-of-the-art machine learning (ML) methods such as backpropagation and reservoir computing recurrent neural networks \cite{vlachas2020backpropagation}, as well as Markov neural operators \cite{likovachki}. 
\end{abstract}


\section{Introduction}

With the continuous advancement of machine learning (ML) methodologies and increasing availability of dedicated software and hardware, data-driven approaches have gained popularity across many different fields such as finance, medicine, science, and engineering. Across these disciplines, one important topic of interest is describing how a dynamical system changes through time to predict its future state \cite{devaney2018introduction}. While some dynamical systems can be predicted through traditional physics-based modeling, there are many real-world examples of systems with unpredictable or chaotic behavior  \cite{hernandez2018chaotic,wang2019neural}. There are many different approaches for predicting and describing chaotic systems such as physics-based modeling, deep learning, physics-informed machine learning, deep operator inference (DeepONets), and operator inference (OpInf).

Physics-based models describing chaotic systems mainly rely on complex partial differential equations (PDEs) that require a tremendous amount of computational power running continuously on high-performance computing (HPC) infrastructure to be solved. Moreover, the systems' non-stationarities, nonlinearities, and intermittency make them intractable from a deterministic standpoint, rendering long-term forecasting unrealistic \cite{wang2019neural}. Reduced-order models (ROMs) aim to solve the computational complexity issue by creating a low-rank representation of the original system that is capable of accurately approximating its full spatiotemporal evolution \cite{brunton2019data,peherstorfer2016data}. This allows significant improvements in computational speed due to the lower number of variables at stake\cite{mohan2018deep,brunton2019data}.

ROMs are typically classified into two types: intrusive and non-intrusive. Intrusive ROMs project the high fidelity full order models (FOM) to a low-dimensional subspace through techniques such as proper orthogonal decomposition (POD) \cite{yildiz2021reduced}. They require access to the differential operators of the FOM to compute the reduced operators. This means that the solution is exclusive to the equation studied, thus, nongeneralizable. Non-intrusive methods have no such drawback; they adopt a data-driven approach where snapshots, i.e. measurements of the states of the dynamical system, are used to learn the ROM \cite{peherstorfer2016data,yildiz2021reduced}. Therefore, non-intrusive approaches can be applied in settings where the FOM operators are unavailable such as in the case of proprietary simulation software \cite{mcquarrie2021data,ghattas2021learning}.

An alternative approach for physics-based ROMs is deep learning methods. These methods circumvent the massive online computational requirements for solving PDEs by leveraging the offline training capacity of artificial neural networks with many hardware architectures such as GPUs, TPUs, CPUs, and cloud services \cite{wang2019neural}. However, one also expects challenges during the offline process such as long training time, availability of RAM, hyperparameter optimization, and scaling for large datasets. Parallelization schemes, such as the one implemented by \cite{vlachas2020backpropagation,pathak2018model}, can mitigate the scaling problem by training multiple models simultaneously. These parallel schemes for ROMs incorporate two extra hyperparameters associated with the training subset sizes and the communication band between neighboring subsets which makes the convergence to an optimal ROM harder. Parallel models are prone to produce less representative ROMs since they strongly depend on additional hyperparameter tuning \cite{vlachas2020backpropagation,nogueira2021reservoir}.

Physics-informed neural networks (PINNs) aim to improve upon traditional neural networks by imposing physical constraints that must be satisfied. These networks are composed of two parts. The first one is a neural network that works as an approximation function that receives time and space coordinates as input providing an approximate solution $\hat{u}$ of a given PDE. The second part takes the object $\hat{u}$ and applies automatic differentiation (AD) over it to compute the residual format of the PDE which, together with its corresponding boundary (BC) and initial conditions (IC), are plugged in the neural network loss function. Thus, a multi-task learning problem is set through a composite loss function that aims to fit any available data while minimizing the residuals of the PDE, BCs, and ICs. The PDE residual, BC, and IC terms in the loss function act as regularization terms discarding unrealistic solutions and consequently constraining the space of admissible solutions to those that adhere to the imposed physical laws \cite{raissi2019physics}.

PINNs have many advantages compared to traditional deep learning: less data is required, a faster training process, and seamless integration of gappy and noisy data \cite{guo2020solving}. The downsides, however, are that its architecture is not generalizable due to the physics embedding being unique to the problem considered, together with convergence rate and computational cost issues \cite{wang2021understanding,wang2021improved}. Recent advances such as the incorporation of Neural Tangent Kernel \cite{wang2021improved} approach, tailored architectures \cite{wang2021improved} and variational formulations \cite{kharazmi2019variational} represent great strides to overcome the stability and generalization issues of PINNs. Those techniques are key enabling features to make PINNs suitable for a wider range of real-world applications.

Deep Operator Networks (DeepONets) are a new class of ML frameworks that can learn nonlinear operator mapping between infinite-dimensional Banach spaces. These frameworks are built upon the generalized universal approximation theorem for operators ensuring that nonlinear operators that map input entities to the corresponding latent space solutions of a given PDE can be learned \cite{lu2021deeponet}. Although they provide a simple and intuitive model architecture, they require large amounts of high-fidelity data to be trained, whereas the learned operator may not be consistent with the underlying physical laws of the system \cite{wang2021learning}. Since the output of DeepONets is differentiable with respect to the input coordinates, they also inherit the basic properties of PINNs. Both can be combined in a single physics-informed DeepONet framework. Such a combination can regularize the target output satisfying physical constraints through its loss function, which can lead to higher data efficiency and improved accuracy \cite{wang2021understanding}.

The non-intrusive Operator inference (OpInf) technique proposed in \cite{peherstorfer2016data} is another particularly suitable approach within the broad class of operator learning methods since it offsets some of the disadvantages of intrusive ROMs and deep learning schemes. The non-intrusive nature of OpInf leads to a more general and explainable model compared to intrusive ROMs and standard deep learning architectures while having much fewer hyperparameters and a quite straightforward training procedure. Such features make the method pretty affordable and efficient since it doesn't rely on minimizing loss functions through gradient descent like DeepONets or other kinds of deep neural networks. The method learns the operators by solving a simple least-squares regression problem (more details in section \ref{sec_opinf}). In the same spirit as OpInf, we can cite Sparse Identification of Nonlinear Dynamics (SINDy \cite{brunton2019data}, which employs an even more generic ansatz for the latent space dynamics, since it can be represented by any kind of nonlinear function, not only polynomials. The SINDy latent space observables are selected from a library of candidates whose coefficients are determined via a sparse regression algorithm (usually using $L_0$ regularization terms). Even though OpInf can be thought of as a subclass of SINDy, the OpInf is generic enough to describe a large class of complex dynamics since any nonlinear representation can be transformed into higher-order polynomials through lifting maps \cite{qian2020lift}.

In a previous work, the authors applied echo-state networks (ESN) alongside ROMs to capture the most relevant features of chaotic dynamical systems and analyzed their forecasting capability \cite{nogueira2021reservoir}. The benchmark chaotic systems Lorenz 63/96 and the GEOS Composition Forecasting (GEOS-CF) dataset for atmospheric pollutants over the continental United States were used as examples. For the Lorenz 63 and 96, the ESN showed remarkable forecasting feature up to 11.8 and 22.8 Lyapunov time units respectively. For the GEOS-CF dataset, ESN showed fair results with pointwise errors of about 15\% for a target 44-hour time horizon. However, that model was unable to perform long-term recursive predictions, being effective only as a single step ahead predictor. In this case, the ROM could reduce the number of dimensions from $24,000$ to $200$ while maintaining 99\% of the system's total energy but showed a limited performance due to a lack of any physical driven mechanism, which turned the extrapolation process of the reduced time series quite hard. Such behavior demonstrated that even in reduced space with $200$ dimensions, a pure data-driven ML approach based on ESN can fail in a long-term extrapolation regime. While traditional ML approaches can deal fairly well with classical chaotic systems, they are still not quite generalizable to more realistic systems.

The learned experience with dense neural networks, LSTMs, and reservoir computing echo-state networks (RC-ESN) empirically evidenced the deficit of robustness, generality, causal structure, and computational efficiency of such algorithms in the long-term extrapolation of chaotic systems. The excessive number of tunable hyperparameters is a common shortage shared by all these Machine Learning (ML) techniques. Only RC-ESN does not suffer from the costly gradient descent training phase. PINNs historically demonstrated issues in simulating chaotic systems due to the lack of a causality mechanism \cite{sifan_shyam}. Likewise, DeepONets showed similar limitations regarding causality while requiring large amounts of paired input-output observations \cite{wang2021learning}. The OpInf, in turn, brings together many desired properties for surrogate modeling, such as causality preserving, computational efficiency, explainable and straightforward linear algebra building blocks, and generalizing skills. It incorporates physical knowledge of the PDEs that govern a broad class of dynamical systems with quadratic nonlinearities such as the Euler equations, incompressible Navier-Stokes equations, shallow water equations, and the Rayleigh-Bénard convection equations, to name a few. All of these PDEs are well represented by a linear-quadratic approximation in latent space upon a Galerkin projection based on a PCA decomposition. The linear-quadratic approximation assumption of OpInf is key to ensuring physical information is embedded into the ROM. The use of lifting maps to transform non-polynomial dynamics into higher-order polynomial dynamics also allows OpInf to be used for any nonlinear PDE \cite{qian2020lift}.

In this work, we apply the OpInf method to approximate the solution of the classical Lorenz 96 and Kuramoto-Sivashinsky chaotic systems, two recurrently studied problems that lay the groundwork for diving into multi-scale fluid flow problems. We adopt OpInf to demonstrate the suitability of physics-inspired ROMs over pure ML-based ROMs to forecast complex spatiotemporal dynamics \cite{vlachas2020backpropagation, pathak2018model}. To verify the robustness of the OpInf approach, we solve both problems over many randomized initial conditions and statistically represent the results. We set up the goal of examining the forecasting capability of the OpInf and show that it is superior to many state-of-the-art black-box ML algorithms in terms of accuracy and efficiency. Using a straightforward multivariate ridge regression technique, we build a computationally parsimonious ROM strategy which proves much cheaper than the established ML algorithms we compare our results with. The OpInf choice also entails environmental responsibility since it commits to a smaller carbon footprint by sparing computational power. The regularization scheme associated with the ridge regression algorithm is vital to avoid overfitting solutions, although quite sensitive to the strategy for finding the best penalization parameters. Even though this work demonstrates that OpInf is quite efficient and robust, there is still much room for improving the OpInf regularization strategies and controlling time integration errors in latent space during the forecasting stage. To the best of our knowledge, this is the first time OpInf is applied to both Lorenz 96 and Kuramoto-Sivashinsky chaotic systems, which we recognize as the preliminary step to tackle more realistic dynamical systems such as atmospheric and oceanic models.

The main contributions of this work are twofold. First, we assess the OpInf skills for improving the predictability of chaotic systems with and without relying on dimensionality reduction, which is illustrated by the Kuramoto-Sivashinsky and Lorenz 96 dynamical systems, respectively. We highlight the merits and caveats of such an algorithm and demonstrate the superiority of the OpInf compared with its main black-box ML competitors such as backpropagation and reservoir computing recurrent neural networks through a thorough statistical analysis. Second, we propose a parallelization scheme to assemble the matrix of physical observables of the OpInf method. The parallelization scheme enables the handling of large spatiotemporal datasets preventing RAM overload while assembling the matrix of observables which quickly scales with the number of latent space dimensions and input snapshots.

The paper is organized as follows: sections \ref{sec:L96} and \ref{sec:KS} state the mathematical description of the chaotic systems considered, namely, Lorenz 96 and Kuramoto-Sivashinsky. Section \ref{sec_opinf} details the Operator Inference approach describing its formulation in matrix notation. Section \ref{sec:model_perform} discusses the metrics and criteria adopted to evaluate model performance. Then, we present numerical results in section \ref{sec:num_res}. Finally, we summarize the main accomplishments of this work, discuss the results, and indicate future research directions in section \ref{sec:conclusions}.

\section{Lorenz 96}\label{sec:L96}

The Lorenz 96 equations are a 3-tier extension of the original Lorenz 63 model, resulting in three coupled nonlinear ODEs. Such system of equations are shown below:

\begin{align}
     \frac{d X_{k}}{d t}  &=X_{k-1}\left(X_{k+1}-X_{k-2}\right)+ F-\frac{h c}{b} \, \Sigma_{j} Y_{j, k} \label{eq:1_Lorenz} \\
\frac{d Y_{j, k}}{d t} &=-c \, b \, Y_{j+1, k}\left(Y_{j+2, k}-Y_{j-1, k}\right)-c \, Y_{j, k} + \frac{h c}{b} \, X_{k}-\frac{h e}{d} \, \Sigma_{i} Z_{i, j, k} \label{eq:2_Lorenz} \\
\frac{d Z_{i, j, k}}{d t} &=e \, d \, Z_{i-1, j, k}\left(Z_{i+1, j, k}-Z_{i-2, j, k}\right)-g \, e \, Z_{i, j, k} + \frac{h e}{d} \, Y_{j, k} \label{eq:3_Lorenz}
\end{align}

This improved model encompasses additional features to its predecessor Lorenz 63 including a large-scale forcing $F$ that makes the system highly chaotic and a set of constant coefficients $b, c, d, e, g, h$ that can be tuned to produce appropriate spatiotemporal variability in all state variables $X_{k},Y_{j, k}$ and $Z_{i, j, k}$. Lorenz 96 was designed to model the large-scale behavior of the mid-latitude atmosphere \cite{chattopadhyay2020data}. The numerical setup of this problem is outlined in section \ref{res:lor}.

\section{Kuramoto–Sivashinsky equation}\label{sec:KS}

The Kuramoto–Sivashinsky (KS) equation was developed almost simultaneously by Yoshiki Kuramoto and Gregory Sivashinsky while studying the turbulent state in a chemical reaction system \cite{kuramoto1978diffusion} and the hydrodynamic instability in laminar flames \cite{sivashinsky1977nonlinear}. This equation can be written as:

\begin{equation}
    \frac{\partial u}{\partial t} + u\frac{\partial u}{\partial x} + a\frac{\partial^2 u}{\partial x^2} + b\frac{\partial^4 u}{\partial x^4} = 0 \label{eq_kuramoto} \\
\end{equation}

The parameters $a$ and $b$ in Equation \ref{eq_kuramoto} were both chosen as unitary. We considered a smooth initial condition for the KS system, which is shown in Equation \eqref{ks_boundary}, and periodic boundary conditions that match the periodicity of the initial condition. The boundary conditions are given by Equations \eqref{periodic1}, \eqref{periodic2}, \eqref{periodic3}, in which we choose the domain size $L=200$, as adopted in \cite{vlachas2020backpropagation}. 
\begin{equation}
    u(0,x) = cos\left(\frac{\pi x}{20}\right)\left[1+sin\left(\frac{\pi x}{20}\right)\right]\label{ks_boundary}
\end{equation}

\begin{equation}
    u(t,0) = u(t,L) \label{periodic1}
\end{equation}

\begin{equation}
    u_x(t,0) = u_x(t,L) \label{periodic2}
\end{equation}

\begin{equation}
    u_{xx}(t,0) = u_{xx}(t,L) \label{periodic3}
\end{equation}

The numerical setup and parameters used in solving Kuramoto-Sivashinsky's equation are detailed in section \ref{res:ks}.

\section{Operator inference}\label{sec_opinf}
Operator inference (OpInf) provides a non-intrusive way of creating ROMs by learning the physical operators of the FOM in latent space by solving a least-squares optimization problem \cite{yildiz2021reduced}. Given measurements of the full order model, an approximate output operator that describes the mapping from the full state to the outputs is found \cite{peherstorfer2016data}. 

Upon discretization, many important dynamical systems including those considered in this work can be cast into a linear-quadratic system of ODEs as follows:

\begin{equation}
\begin{split}
    \frac{d}{dt}\mathbf{q}(t) =\mathbf{c}+\mathbf{Aq}(t)+\mathbf{H}(\mathbf{q}(t)\otimes \mathbf{q}(t)) + \mathbf{Bu}(t), \\
    \mathbf{q}(t_0) = \mathbf{q_0}, t \in [t_0,t_f] \label{eq:1}
\end{split}
\end{equation}

where $\mathbf{q}(t) \in \mathbb{R}^n$ is the state vector at time $t$ with dimension $n$, $\mathbf{q}_0 \in \ \mathbb{R}^n$ is the initial condition, $\mathbf{A} \in \mathbb{R}^{n \times n}$ is the linear ODE operator, $\mathbf{H} \in \mathbb{R}^{n \times n^2}$ is the quadratic ODE operator, $\mathbf{c} \in \mathbb{R}^{n}$ are constant terms, $\mathbf{u}(t) \in \mathbb{R}^{m}$ are input terms related to BCs or forcing terms, $\mathbf{B} \in \mathbb{R}^{n \times m}$ is a linear operator in the forcing term $\mathbf{u(t)}$, and $\mathbf{q} \otimes \mathbf{q} = [q_1^2,q_1 q_2,...,q_1q_n,q_2q_1,q_2^2,...,q_2q_n,...,q_n^2] \in  \mathbb{R}^{n^2}$.

Note that Equation \eqref{eq:1} has a polynomial structure. OpInf is especially suited for dynamical systems with polynomial nonlinearities such as Lorenz 96 and KS. For systems with non-polynomial nonlinear operators, one can use lifting maps to transform non-polynomial dynamics into higher-order polynomial dynamics \cite{peherstorfer2016data,qian2020lift}.

The OpInf method is built upon a sequence of steps starting by gathering a snapshot matrix $\mathbf{Q}$ that contains a finite set of measurements of the FOM, as seen in the following:

\begin{equation}
    \mathbf{Q} = [\mathbf{q}(t_0),\mathbf{q}(t_1),...,\mathbf{q}(t_{f})] \ \in \ \mathbb{R}^{n \times k} \label{eq:2}
\end{equation}
where $k$ is the total number of snapshots.

In the next step, one chooses a low-dimensional basis $\mathbf{V} \in \mathbb{R}^{n \times r}$, with $r \ll n$, to represent each vector $\mathbf{q}(t)$ of the matrix $\mathbf{Q}$. In this case, $\mathbf{q}(t) \approx \mathbf{V} \mathbf{q}_{r}(t)$, where $\mathbf{q}_{r}(t) \in \mathbb{R}^{r}$ is the reduced state vector. This study applied Principal Component Analysis (PCA) to generate the basis $\mathbf{V}$.

A Galerkin projection on Equation \eqref{eq:1} using the basis $\mathbf{V}$ results in a system of ODEs representing the original dynamics in the reduced order space. If the PDE operators of the FOM are known, it is straightforward to check that the Galerkin projection preserves the polynomial structure of the operators. This observation suggests that even if we do not have access to the FOM operators, we can assume that the reduced model operators have the same shape (i.e., the same polynomial structure) as those in Equation \eqref{eq:1}. Such an assumption is central in the OpInf method and constitutes the third building block of the OpInf ROM. Rewriting the projected snapshots $\mathbf{q}_{r}(t)$ as $\widehat{\mathbf{q}}(t)$, the OpInf seeks reduced operators $\widehat{\mathbf{c}} \sim \mathbf{c}$, $\widehat{\mathbf{A}} \sim \mathbf{A}$, $\widehat{\mathbf{H}} \sim \mathbf{H}$, $\widehat{\mathbf{B}} \sim \mathbf{B}$ that mimic the original full order operators in a non intrusive way, which are shown in Equation \eqref{eq:3} 
\begin{equation}
\begin{split}
    \frac{d}{dt}\mathbf{\widehat{q}}(t) =\mathbf{\widehat{c}}+\mathbf{\widehat{A}\widehat{q}}(t)+\mathbf{\widehat{H}}(\mathbf{\widehat{q}}(t)\otimes \mathbf{\widehat{q}}(t)) + \mathbf{\widehat{B}u}(t), \\
    \mathbf{\widehat{q}}(t_0) = \mathbf{V}^{T}\mathbf{q_0}, t \in [t_0,t_f] \label{eq:3}
\end{split}
\end{equation}

Given the projected state variables $\mathbf{\widehat{q}}(t)$ and its time derivatives $\mathbf{\dot{\widehat{q}}}(t)$, calculated with any numerical procedure such as finite differences methods and spline interpolation, the best possible match of the reduced operators with respect to the reduced-order system of ODEs can be found by minimizing the residual of Equation \eqref{eq:3}. Thus, the fourth and final building block of the OpInf is to solve a least-squares problem, where the objective function seeks to minimize the residual of the latent dynamics in the Euclidean norm and the objective variables are the reduced operators $\widehat{\mathbf{c}}, \widehat{\mathbf{A}}, \widehat{\mathbf{H}}, \widehat{\mathbf{B}}$. This problem is ill-conditioned and prone to overfitting \cite{mcquarrie2021data}. To mitigate this issue, we can rewrite the least-squares problem in a regularized version as shown below \eqref{eq:5}.
\begin{equation}
\begin{split}
\min _{\widehat{c}, \widehat{A}, \widehat{\mathbf{H}}, \widehat{\mathbf{B}}} \sum_{j=0}^{k-1}\left\|\widehat{\mathbf{c}}+\widehat{\mathbf{A}} \widehat{\mathbf{q}}_{j}+\widehat{\mathbf{H}}\left(\widehat{\mathbf{q}}_{j} \otimes \widehat{\mathbf{q}}_{j}\right)+\widehat{\mathbf{B}} \mathbf{u}_{j}-\dot{\widehat{\mathbf{q}}}\right\|_{2}^{2}+ \\ \lambda_{1}\|\widehat{\mathbf{c}}\|_{2}^{2}+\lambda_{2}\|\widehat{\mathbf{A}}\|_{F}^{2}+\lambda_{3}\|\widehat{\mathbf{H}}\|_{F}^{2}+\lambda_{4}\|\widehat{\mathbf{B}}\|_{F}^{2} \label{eq:5}
\end{split}
\end{equation}

For the sake of conciseness, the regularized regression problem can be cast into a matrix form written as 

\begin{equation}
\min _{\mathbf{O}}\left\|\mathbf{D O}^{\top}-\mathbf{R}^{\top}\right\|_{F}^{2}+\left\|\mathbf{\Gamma O}^{\top}\right\|_{F}^{2} \label{eq:6}
\end{equation}

where
\begin{equation}\label{eq:opinf_detail}
\begin{aligned}
&\mathbf{O}=\left[\begin{array}{llll}
\widehat{\mathbf{c}} & \widehat{\mathbf{A}} & \widehat{\mathbf{H}} & \widehat{\mathbf{B}}
\end{array}\right] \in \mathbb{R}^{r \times d(r, m)} \\
&\mathbf{D}=\left[\begin{array}{llll}
\mathbf{1}_{k} & \widehat{\mathbf{Q}}^{\top} & (\widehat{\mathbf{Q}} \otimes \widehat{\mathbf{Q}})^{\top} & \mathbf{U}^{\top}
\end{array}\right] \in \mathbb{R}^{k \times d(r, m)} \\
&\widehat{\mathbf{Q}}=\left[\begin{array}{llll}
\widehat{\mathbf{q}}_{0} & \widehat{\mathbf{q}}_{1} & \cdots & \widehat{\mathbf{q}}_{k-1}
\end{array}\right] \in \mathbb{R}^{r \times k} \\
&\mathbf{R}=\left[\begin{array}{llll}
\dot{\widehat{\mathbf{q}}}_{0} & \dot{\widehat{\mathbf{q}}}_{1} & \cdots & \dot{\widehat{\mathbf{q}}}_{k-1}
\end{array}\right] \in \mathbb{R}^{r \times k} \\
&\mathbf{U}=\left[\begin{array}{llll}
\mathbf{u}_{0} & \mathbf{u}_{1} & \cdots & \mathbf{u}_{k-1}
\end{array}\right] \in \mathbb{R}^{m \times k}  \\
&\mathbf{\Gamma}= \lambda \mathbf{I} = \Gamma(\lambda_1,\lambda_2,\lambda_3,\lambda_4), \ \lambda > 0 .
\end{aligned} 
\end{equation}

In this set of equations $d(r,m) = 1 + r + \binom{r+1}{2} + m$ and $\mathbf{1}_{k} \in \mathbb{R}^{k}$ is a column vector of length $k$ with unity entries.
We notice that in Equation \eqref{eq:6}, $\mathbf{O}$ contain the unknown operators, $\mathbf{D}$ are known data values,  $\widehat{\mathbf{Q}}$ are the projected state variables, $\mathbf{R}$ are the time derivatives of the projected state variables, $\mathbf{U}$ represents possible forcing terms, and $\mathbf{\Gamma}$ is a diagonal regularizer, whose entries $\lambda_1,\ldots,\lambda_4$ are defined according to the rule showed in Appendix \ref{lambda_gamma}. For the sake of simplicity, we fixed $\lambda_1 = \lambda_2$ and set $\lambda_4 = 0$ since the numerical examples considered in this work use no forcing term.

The minimizer of the regression problem in Equation \eqref{eq:6} satisfies the modified normal equations:

\begin{equation}
\left(\mathbf{D}^{\top} \mathbf{D}+\mathbf{\Gamma}^{\top} \boldsymbol{\Gamma}\right) \mathbf{O}^{\top}=\mathbf{D}^{\top} \mathbf{R}^{\top} \label{eq:8}
\end{equation}

which constitutes an algebraic linear system that can be solved by a variety of efficient algorithms. Furthermore, the optimization problem could also be transformed into $n$ independent least-squares problems and solved efficiently using standard solvers in parallel architectures \cite{geelen}. 

The most computationally expensive operation in the closed-form construction stage (cf. Equation \eqref{eq:8}) is the evaluation of $\D^T\,\D$. Matrix $\D$ is usually dense and, depending on the number of samples and degrees of freedom required to fit the model, it becomes challenging to allocate such matrix operations to a single computational node. For instance, this work's snapshot matrix $\mathbf{Q}$ of the KS test case is a $480,000$ by $160$ matrix representing $586$ MB of storage. However, the assembling of matrix $\mathbf{D}$ (Eqs. \ref{eq:opinf_detail} and \ref{eq:8}) produces more than $47$ GB of storage. Such operation illustrates how memory requirements can quickly scale as the input data increases. By splitting the inner product operation into batch-wise operations along the samples axis and dispatching them in multiple MPI processes, we can rely on huge amounts of data for creating $\D$ without suffering memory issues. Appendix \ref{parallel_eval} shows a pseudocode detailing this batch-wise approach. Alternatively, one can compute a solution for the algebraic linear system by using the Moore-Penrose pseudoinverse $\mathbf{O}^{\top}=\D^\dagger\mathbf{R}^{\top}$ in which no regularization parameter is necessary. 

When OpInf learns the reduced operators, it indirectly considers information coming from the original PDE and, thus, learns the underlying physics of the dynamical system without having access to the full order operators showed in Equation \eqref{eq:1}. The incorporation of physics principles into the machine learning pipeline is the main motivation for choosing OpInf as a chaotic dynamical system forecasting tool.

\section{Quantifying model performance}\label{sec:model_perform}

To quantify a system's chaoticity, it is common to study its sensitivity to initial conditions. Consider two initial conditions represented by the vectors $x(t)$ and $x(t)+\delta(t)$, where $\delta(t)$ is an infinitesimally small displacement. At time $t=0$, the distance between both trajectories is set as  $\delta(t=0)=\delta_0$. For chaotic systems, trajectories start diverging exponentially in time. Such divergence can be measured through the following mathematical relation:

\begin{equation}
    \|\delta(t)\| \approx  \|\delta_0\| \  e^{\Lambda t} \label{1}
\end{equation}

The variable $\Lambda$ in Equation \eqref{1} is referred to as the Lyapunov exponent and it characterizes the stability of the system. If the Lyapunov exponent is positive, it means the system is chaotic while negative values represent stability. 

Considering multi-dimensional systems, typically, there are as many Lyapunov exponents as there are dimensions. If at least one positive Lyapunov exponent exists, the system is chaotic, however, the rate of divergence can be qualitatively observed by the magnitude of these exponents. The value of $\Lambda$ can also be different depending on the orientation of the displacement vector $\delta(t)$, given by $\delta(t)/|\delta(t)|$ \cite{vlachas2020backpropagation,wang2019neural}. Therefore, the possible values of $\Lambda$ defines a spectrum $\Lambda_1 \geq \Lambda_2 \geq \Lambda_3 \geq ... \geq \Lambda_n $ with $n$ being the dimensionality of the phase space. Note that when $\Lambda$'s with positive values are dominant, we observe faster divergence among trajectories. 

From Equation \eqref{1}, the largest positive value of $\Lambda$ causes the largest divergence, which is called the Maximal Lyapunov exponent (MLE) denoted by $\Lambda_1$. The MLE has two main purposes:

\begin{itemize}
    \item To measure the level of unpredictability of the dynamical system \cite{wang2019neural};
    \item Provide a characteristic time scale to quantify the quality of predictions based on the forecasting error growth (Lyapunov time) \cite{vlachas2020backpropagation}.
\end{itemize}

In this work, we scale the time axis of all plots in numerical results by the largest Lyapunov exponent $\Lambda_1$ in order to characterize the chaoticity of the system. This is done by multiplying the time vector $t$ by the MLE of the model ($T^{\Lambda_1} = t \Lambda_1$). Aiming at establishing a straightforward comparison with methods based on backpropagation and reservoir computing RNN architectures showed in \cite{vlachas2020backpropagation}, we considered the same setups and geometries used in that paper. Thus, we considered the Lyapunov exponent of the Kuramoto-Sivashinsky problem as $\Lambda_1 = 0.094$ while for Lorenz-96 system we fixed $\Lambda_1=1.68$ for $F=8$ and $\Lambda_1=2.27$ for $F=10$.

To allow direct comparisons with \cite{vlachas2020backpropagation}, this work also considers the valid prediction time $\mathrm{(VPT)}$ to quantify the model's predictive performance through a single metric. The VPT is computed in terms of the system's MLE ($\Lambda_1$) using the definition
\begin{equation}
\mathrm{VPT}=\frac{1}{T^{\Lambda_{1}}} \underset{t_{f}}{\operatorname{argmax}}\left\{t_{f} \mid \operatorname{NRMSE}\left(\boldsymbol{o}_{t}\right)<\epsilon, \forall t \leq t_{f}\right\}, \label{2}
\end{equation}
where $\boldsymbol{o}_{t} \in \mathbb{R}^{n \times m }$ represents a generic set of time-series. Textually, $\mathrm{VPT}$ finds the largest time $t_f$ that the model can forecast with a normalized root mean square error (NRMSE) of the observable time-series $o_t$ smaller than a threshold $\epsilon$ and normalizes it using $\Lambda_1$. The value of $\epsilon$ is set to 0.5 to allow comparisons with \cite{vlachas2020backpropagation}.
$\mathrm{NRMSE}$ is computed as follows
\begin{equation}
\label{eq:nrmse}
    \mathrm{NRMSE}(\mathbf{\widehat{x}}) = \sqrt{\frac{\sum^{N-1}_{i=0} (\mathbf{x}_{i} - \mathbf{\widehat{x}}_{i})^2}{N\boldsymbol{\sigma}^2}} ,  
\end{equation}
where $\mathbf{x}_{i}$ and $\mathbf{\widehat{x}}_{i}$ identify the reference and predicted time-series, respectively, $N$ is the total number of time-series considered (possibly in latent space), and $\boldsymbol{\sigma}$ is the array of standard deviations over time for all time-series. The division in Equation \ref{eq:nrmse} should be understood as an element-wise operation, that is, for each time-series, there is a single $\sigma_{i}$. 

\begin{figure}
\centering
\includegraphics[scale=0.33]{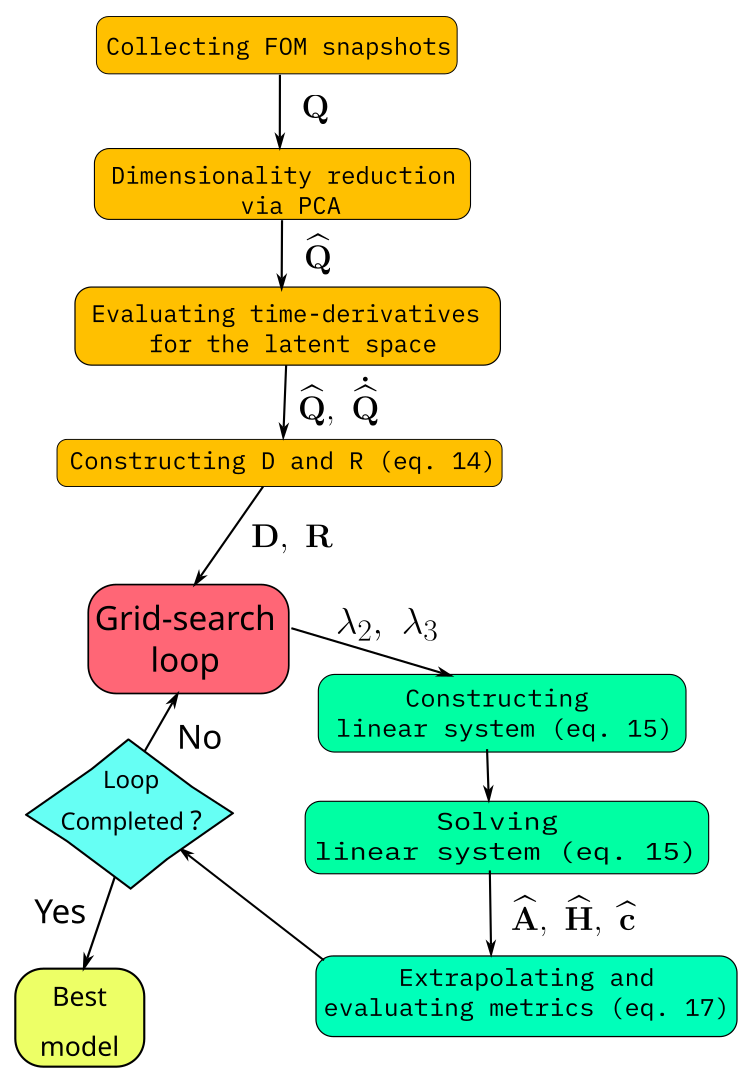}
\caption{OpInf hyper-optimization workflow schematic.}
\label{opinf_schematic}
\end{figure}

Fig. \ref{opinf_schematic} shows a schematic of the whole hyper-optimization workflow used in the OpInf algorithm including the matrix assembling descriptions of the previous sections and the performance metrics evaluation detailed in this section.

\section{Numerical results}\label{sec:num_res}
In this section, we assess the effectiveness of the OpInf method in creating a reduced-order model through two benchmark examples: the Lorenz 96 and the Kuramoto-Sivashinsky systems. The former is a suitable emulator for complex multi-scale atmospheric systems. The latter shows rich dynamical characteristics with many bifurcations depending on the size $L$ of the spatial domain culminating in strong chaotic regimes.

\subsection{Lorenz 96} \label{res:lor}
The Lorenz 96 numerical experiment is performed using $N=40$ degrees of freedom for the interval $t = [0, 2000]\,s$ with a timestep $dt=0.01\,s$. The forcing term $F$ is chosen as either $F=8$ or $F=10$. 

Assuming periodic boundary conditions and starting from a randomly generated initial condition, the problem is solved by employing SciPy's LSODA algorithm, a solver that automatically switches between a nonstiff Adams solver and a Backward Differentiation Formula method for stiff problems. The first $1000\,s$ are discarded to get rid of the initial transient behavior, and the remaining $1000\,s$ are split into two sets of equal size corresponding to the training and testing datasets. Since the Lorenz 96 system is randomly initialized, the simulation is repeated $100$ times for each forcing term to analyze the sensitivity of the reduced model with respect to the randomized ICs. The OpInf reduced models are obtained by solving a least-squares problem via Moore-Penrose pseudoinverse for each random initialization and subsequently used for computing extrapolations. To evaluate the quality of each model, we employ the VPT criteria. For each simulation set, corresponding to a different forcing term $F$, the VPT values are recorded and their minimum, maximum, mean, and standard deviation values are showed in Table \ref{tab:VPT}. In this table $\sigma^{*}$ is computed over all approximate solutions associated to each IC. The VPT threshold value is fixed as $0.5$.

\begin{table}[h!]
\caption{Minimum, Maximum, Avegare and Standard Deviation of VPT for the Lorenz 96 simulations.}
    \centering
    \begin{tabular}{ccc}
     \hline
     $F$ & 8 & 10 \\
     \hline
     VPT min &  7.40 & 7.19 \\ 
     VPT max & 19.15 & 15.39 \\ 
     VPT avg & 12.49 & 12.00 \\ 
     VPT $\sigma^{*}$ & 1.98 & 1.60 \\ 
     \hline
    \end{tabular}
    
    \label{tab:VPT}
\end{table}

\begin{figure}[!h]
\centering
\begin{tabular}{cc}
\includegraphics[scale=0.66]{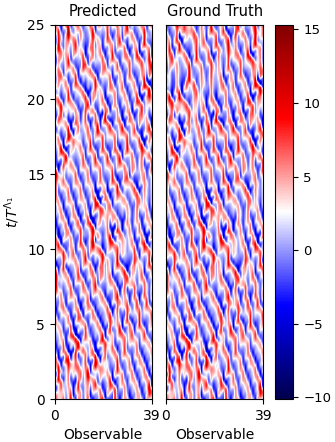} &
\includegraphics[scale=0.66]{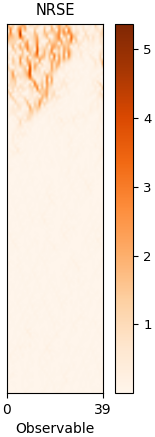} \\
a) & \hspace{-0.6cm} b)
\end{tabular}
\caption{Contour plots of the spatiotemporal forecast of the Lorenz 96 system with $40$ degrees of freedom and $F=8$: a) OpInf prediction vs. ground truth; b) NRSE error. Color bars indicate absolute values: a) system state variables; b) divergence from ground truth (e.g., $1.0$ indicates $100\%$ deviation).}
\label{contour_lorenz_96_F=8}
\end{figure}

\begin{figure}[!h]
\centering
\begin{tabular}{cc}
\includegraphics[scale=0.66]{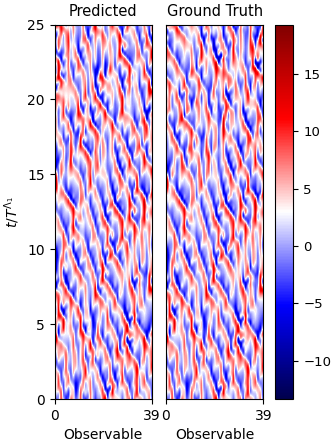} &
\includegraphics[scale=0.66]{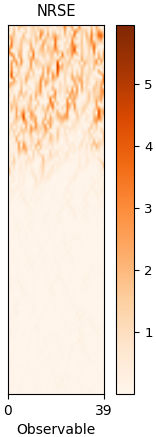} \\
a) & \hspace{-0.6cm} b)
\end{tabular}
\caption{Contour plots of the spatiotemporal forecast of the Lorenz 96 system with $40$ degrees of freedom and $F=10$: a) OpInf prediction vs. ground truth; b) NRSE error. Color bars indicate absolute values: a) system state variables; b) divergence from ground truth (e.g., $1.0$ indicates $100\%$ deviation).}
\label{contour_lorenz_96_F=10}
\end{figure}

The left-hand side of Figs. \ref{contour_lorenz_96_F=8} and \ref{contour_lorenz_96_F=10} compares the approximate and target contour plots of the spatiotemporal forecast of the Lorenz 96 chaotic system for the configurations $F=8$ and $F=10$, respectively. The right-hand side of the same Figs. shows the normalized root square error (NRSE) associated to each one of the forcing configurations. The OpInf results correspond to the best approximation among all random ICs.

\begin{figure}[!h]
\centering
\begin{tabular}{cc}
\hspace{-0.7cm}
\includegraphics[scale=0.48]{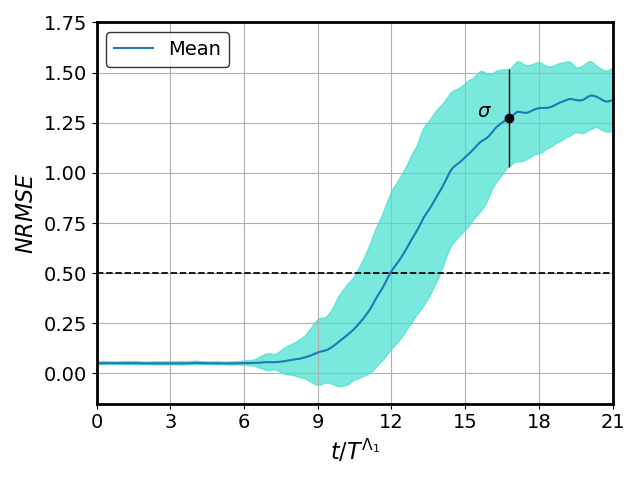} &
\includegraphics[scale=0.48]{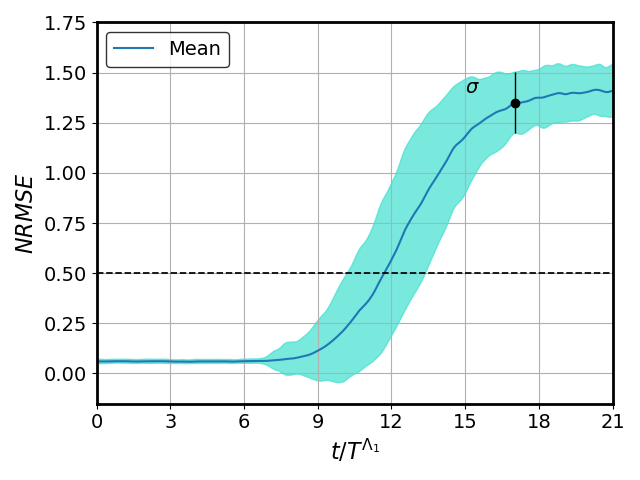} \\
\hspace{0.2cm} a) & \hspace{0.8cm} b)
\end{tabular}
\caption{Mean $NRMSE$ (blue line) and standard deviation envelope for the Lorenz 96 system over 100 random initial conditions sampled from the testing data: a) $F=8$; b) $F=10$. The dashed line indicates the threshold value of the VPT quality metric.}
\label{NRMSE_F=8_F=10}
\end{figure}

Fig. \ref{NRMSE_F=8_F=10} shows the evolution of the mean $NRMSE$ along time for both configurations of the forcing term, i.e., $F=8$ (Fig. \ref{NRMSE_F=8_F=10}a) and $F=10$ (Fig. \ref{NRMSE_F=8_F=10}b) considering all random ICs. The plots also depict the standard deviation envelope around the average $NRMSE$ which is computed as $\overline{NRMSE} \pm \sigma(NRMSE)$. These outcomes indicate a clear prediction improvement of the OpInf method over the backpropagation and reservoir computing RNN architectures considered in \cite{vlachas2020backpropagation}. For instance, comparing the average $NRMSE$ for F=8 and F=10 reported in \cite{vlachas2020backpropagation} with the ones in Fig. \ref{NRMSE_F=8_F=10}, their best forecasting model reached VPT at $0.79$ and $0.83$ Lyapunov time units (cf. Table 1 in \cite{vlachas2020backpropagation}) respectively, while OpInf did the same at $12.49$ and $12.00$ Lyapunov time units (Table \ref{tab:VPT}). Note that if we consider the minimum VPT values in the Table \ref{tab:VPT}, $7.40$ and $7.19$, for $F=8$ and $F = 10$, they are both superior to the best $NRMSE$ in \cite{vlachas2020backpropagation} which equal to $2.31$ and $2.35$, respectively. 

Fig. \ref{lorenz_96_modes_F=8,10} shows the predictive power of OpInf for five different discrete variables of the Lorenz 96 model with $F=8$ and $F=10$, respectively. The plots compare the approximate integrated solution with the ground truth over time for the best approximate solution among all random ICs. We can observe from Figs. \ref{contour_lorenz_96_F=8}, \ref{contour_lorenz_96_F=10},  and \ref{lorenz_96_modes_F=8,10} that the OpInf performed remarkably for both $F=8$ and $F=10$ configurations matching the exact solutions for as long as $\approx 18$ and $\approx 15$ Lyapunov time units, respectively.

\begin{figure}[!h]
\centering
\begin{tabular}{cc}
\includegraphics[scale=0.35]{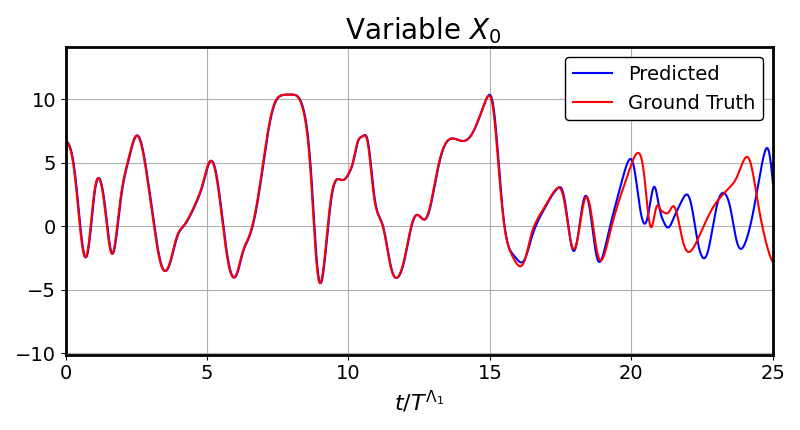} & \includegraphics[scale=0.35]{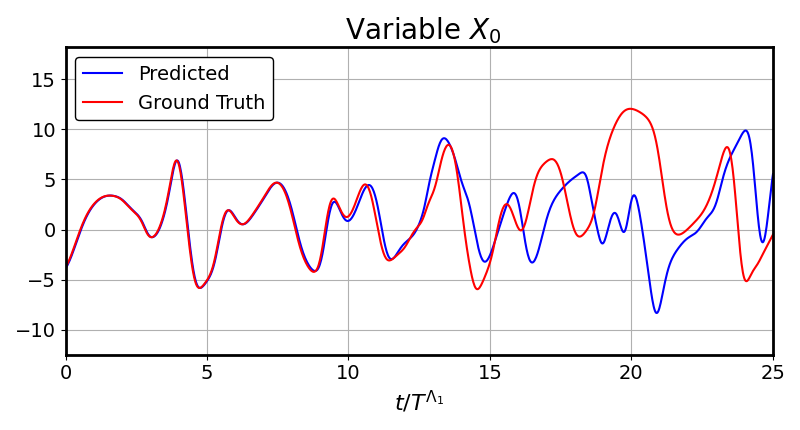} \\
\includegraphics[scale=0.35]{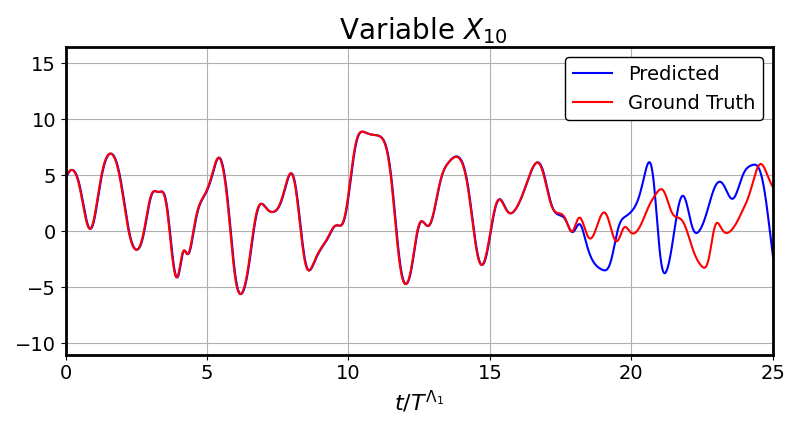} &
\includegraphics[scale=0.35]{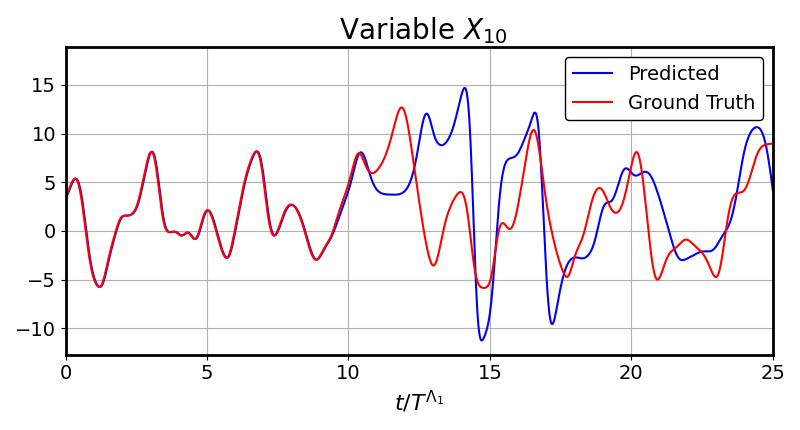} \\
\includegraphics[scale=0.35]{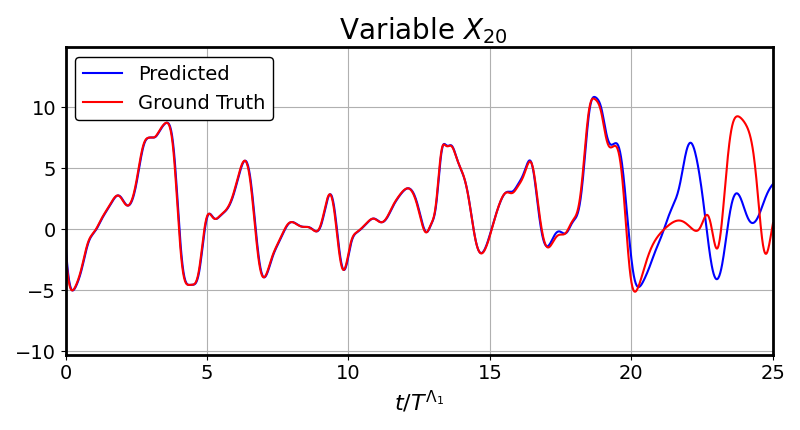} &
\includegraphics[scale=0.35]{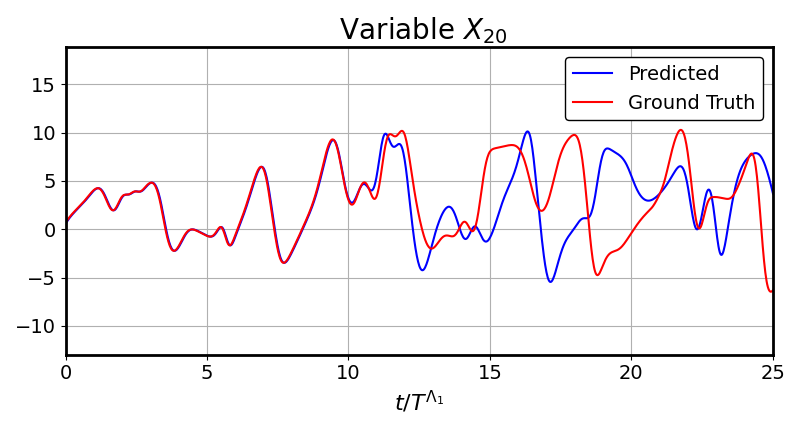} \\
\includegraphics[scale=0.35]{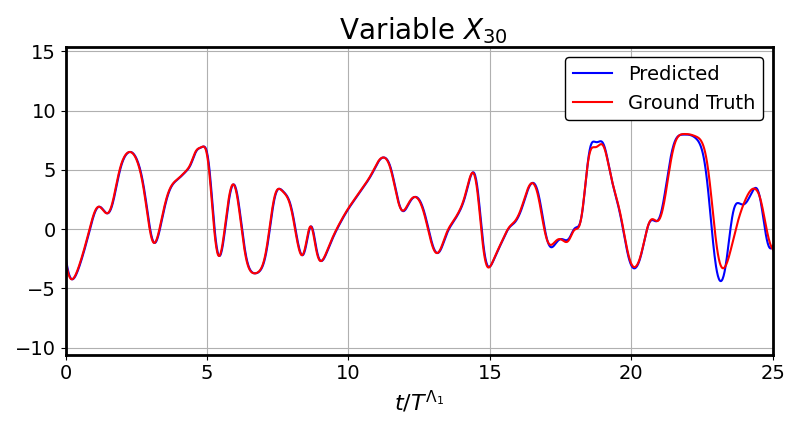} &
\includegraphics[scale=0.35]{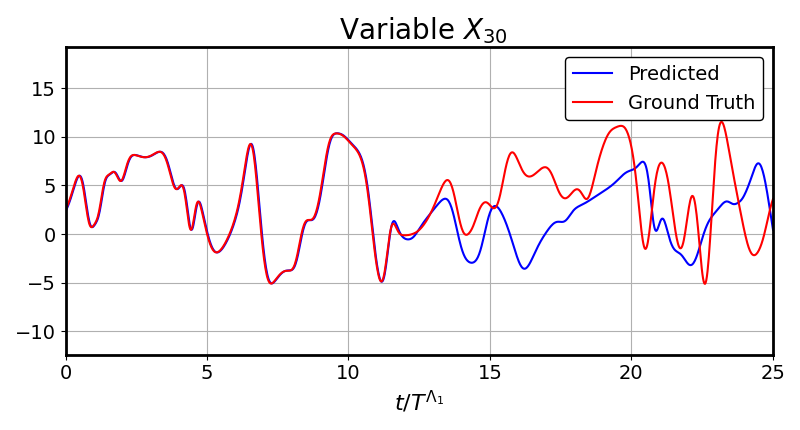} \\
\includegraphics[scale=0.35]{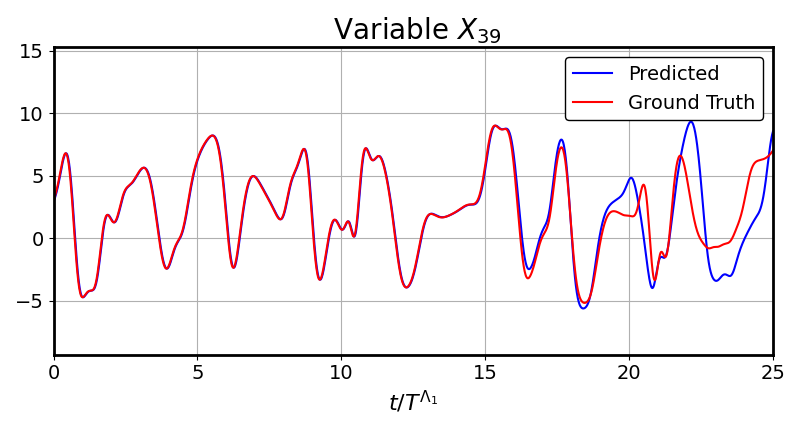} &
\includegraphics[scale=0.35]{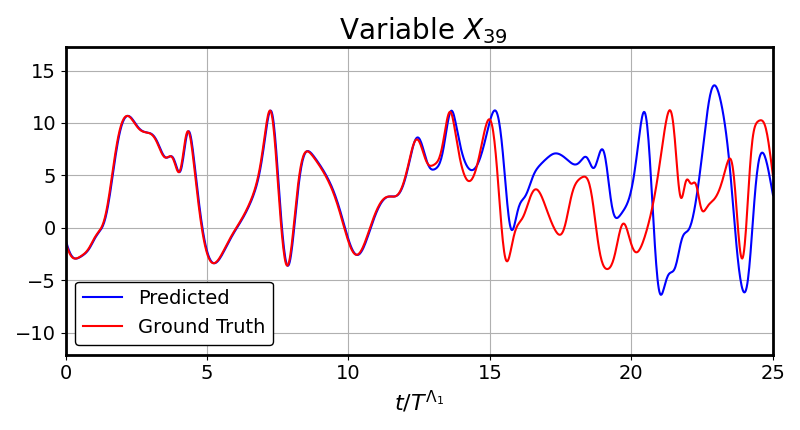} \\
\hspace{0.1cm} a) & \hspace{0.2cm} b)
\end{tabular}
\caption{Predicted (blue) vs. ground truth (red) integrated discrete variables $X_{k}, \, k = 0,10,20,30,39$ for the Lorenz 96 system: a) $F=8$; b) $F=10$.}
\label{lorenz_96_modes_F=8,10}
\end{figure}

Regarding the computational efficiency aspect of the OpInf ROM, we observe that the training time of a single realization for the Lorenz 96 system (i.e., for a unique choice of hyperparameters) takes $3.8\,s$ to deliver an output using the Moore-Penrose pseudoinverse algorithm. The results presented in \cite{vlachas2020backpropagation} report more than $6$ GPU hours using $12,000$ processing nodes for the parallel reservoir computing scheme to do the same job. For LSTM architectures, they report more than $7$ GPU hours to accomplish the same task. All the numerical experiments in this work were executed in a 20-core Intel Xeon CPU.

\subsection{Kuramoto-Sivashinsky}\label{res:ks}

The Kuramoto-Sivashinsky (KS) equation is defined over a domain $\Omega = [0,L]$ with periodic boundary conditions. We chose the domain size $L = 200 \,m$ and discretized it with $512$ equally spaced grid points. The KS equation marches in time for a total $T = 6 \cdot 10^4 \,s$ with a time step $dt = 0.125\,s$ corresponding to $48 \cdot 10^4$ steps. The reference solution of the KS equation is generated using the modified exponential time-differencing fourth-order Runge-Kutta scheme (ETDRK4) as described in \cite{Kassam2005}. The ETDRK4 scheme combines a modified version of the exponential time-differencing (ETD) scheme with the fourth-order Runge-Kutta (RK4) time integrator.  Time-stepping is performed via a fourth-order Runge-Kutta method using complex analysis in which each function evaluation is performed using contour integrals in the complex plane. This technique helps alleviate numerical difficulties.

\begin{figure}[!h]
\centering
\includegraphics[scale=0.46]{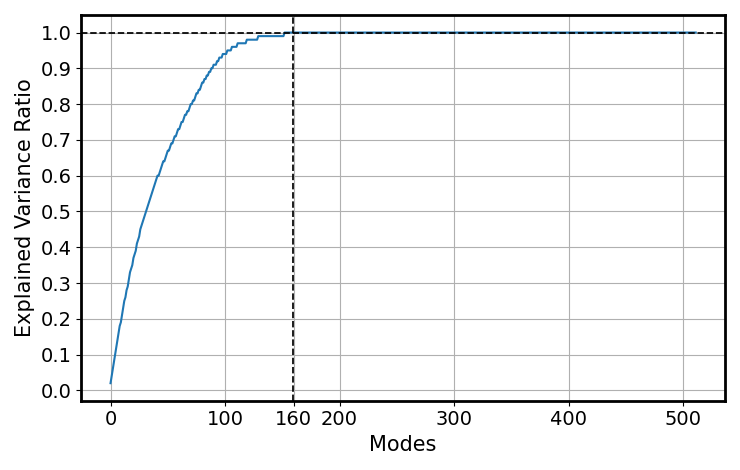}
\caption{Cumulative explained variance ratio for the PCA decomposition of the Kuramoto-Sivashinsky dataset. The vertical dashed line indicates the total number of eigenmodes used by the OpInf ROM which corresponds to $99.99\%$ explained variance.}
\label{cumul_energy_PCA}
\end{figure}

We employ the Principal Component Analysis (PCA) to reduce the dimensionality of the full order model from $512$, which corresponds to the number of grid points, to $r=160$ dimensions that match the number of reduced basis vectors in $\mathbf{V}$. The final shape of the spatiotemporal dataset used to train and test the OpInf model is set to $(t,x) = (4,8\,\times 10^5, 160)$. The PCA decomposition is applied to $90\%$ of the initial $(t,x)$ snapshots and the choice of $k$ is guided by the projection error evaluated over the remaining $10\%$ of the dataset. The projection error obtained in this region is approximately $0.51 \%$ and the expected variance ratio higher than $0.9999$ with respect to the $90\%$ portion of the initial snapshots. Fig. \ref{cumul_energy_PCA} shows the cumulative explained variance ratio (or the energy content) preserved by the PCA decomposition as a function of the number of modes.

In order to further justify the previous choice regarding the number of PCA modes $r$ that provide a good approximation capability for the OpInf model, apart from the energy content preservation with respect to the full order model, we assessed the impact of an increasing number of modes on the VPT criterion as shown in Table \ref{tab:vpt_r}. In this table, VPTs are evaluated at the beginning of the validation dataset. This result demonstrates that a large enough reduced space is necessary to produce reliable approximations in the full space, and that is another strong reason for fixing $r = 160$. We also notice that the VPT values in Table \ref{tab:vpt_r} do not reflect the largest possible VPTs for the best models since it is applied for a single initial condition (IC).

\begin{table}
	\centering
	\caption{VPT as a function of the PCA modes $r$.}
\begin{tabular}{c c c c c c}
	\hline
	$r$ & 20 & 55 & 100 & 120 & 160 \\
	\hline
	VPT & 0 & 0.094 & 1.175 & 2.385 & 4.125  \\
	\hline
\end{tabular}
\label{tab:vpt_r}
\end{table}

The modeling dataset is split into three subsets with proportions (90\%, 5\%, 5\%) corresponding to the training, validation and test datasets, respectively. The OpInf reduced model is generated considering a simple grid-search for the $log_{10}$ of the Thikonov regularization penalties $(\lambda_2, \lambda_3)$ for the discrete domain $\left\{(\lambda_2,\lambda_3): -3 \leq log_{10} \,\lambda_2 \leq 3, -3 \leq log_{10}\,\lambda_3 \leq 3\right\}$ with a $0.5$ step for both $\lambda_2$ and $\lambda_3$. For this particular case, we found the combination $(\lambda_2,\lambda_3) = (10^{-3}, 10^{-3})$ as the best option for minimizing the extrapolation error in the validation dataset. Furthermore, the regularizers $\lambda_1$ and $\lambda_2$ are chosen to be equal, and $\lambda_4$ is set to zero since the constant polynomial term $\mathbf{\widehat{c}}$ (cf. Equation \eqref{eq:3}) has minor influence over the final results and the KS system has no forcing term to fit the operator $\mathbf{\widehat{B}}$. Once OpInf operators are obtained, they are used for extrapolating the state variable $\widehat{\mathbf{q}}(t)$ over an unseen dataset. In order to evaluate the model sensibility to different initial conditions, we proceed by time-integrating the reduced model with the SciPy routine \textit{odeint} using a set of 100 initial conditions randomly selected from the test dataset. We use the same time horizon of 90 s (empirically chosen) to assess the model extrapolation for each IC. Table \ref{tab:VPT_ks} shows minimum, maximum, mean, and standard deviation VPT values for the KS system. In this table, $\sigma^{*}$ is computed over all approximate solutions associated to each IC. The VPT threshold in this case is fixed as $0.5$.

\begin{table}[h!]
\caption{Minimum, Maximum, Average and Standard Deviation of VPT for the Kuramoto-Sivashinsky simulations.}
    \centering
    \begin{tabular}{ccc}
     \hline
     VPT min &  2.53 \\ 
     VPT max &  7.01 \\ 
     VPT avg &  4.60 \\ 
     VPT $\sigma^{*}$ & 0.95 \\
     \hline
    \end{tabular}
    
    \label{tab:VPT_ks}
\end{table}

\begin{figure}[!h]
\centering
\includegraphics[scale=0.48]{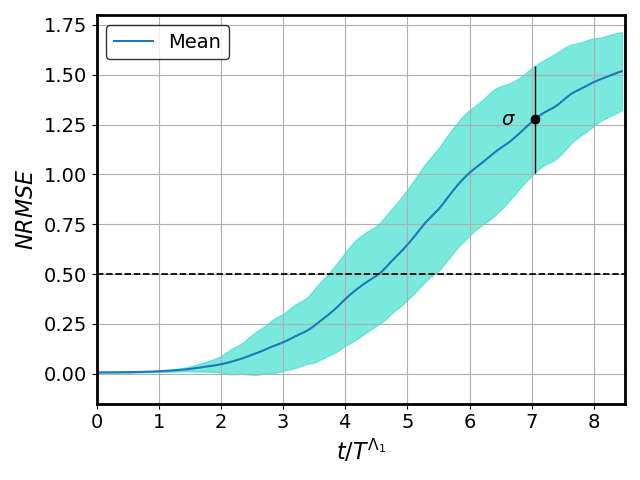}
\caption{Mean $NRMSE$ (blue line) and standard deviation envelope for the Kuramoto-Sivashinsky case over 100 initial conditions sampled from the testing data. The dashed line indicates the threshold value of the VPT quality metric.}
\label{NRMSE_KS}
\end{figure}

The extrapolated solutions are then reconstructed into the full dimensional space through the PCA basis $\mathbf{V}$. The relative approximation error in $L^2$-norm achieved by the best reconstructed solution among all time-integrated solutions associated to each IC is $25.8 \%$ for the time horizon of $90 s$. Considering only the first $75 s$ in the test dataset comprising the $90 s$, the relative approximation error drops to $1.2 \%$. The VPT metric associated to the best case achieves the significant value of $7.01$. The NRMSE is evaluated for the $160$ latent time-series $\mathbf{\widehat{x}}_{i}$ referenced in Equation \ref{eq:nrmse}. A summary of the VPT results for this numerical experiment can be seen in Figure \ref{NRMSE_KS}. Comparing OpInf performance for the KS system with the results in \cite{vlachas2020backpropagation}, OpInf is $43\%$ superior in terms of forecasting capability. We should also notice that the OpInf performed remarkably well with a reduced number of dimensions (i.e., using $160$ eigenmodes). In contrast, all the network architectures analyzed in \cite{vlachas2020backpropagation} produced forecasts \textcolor{blue}{only} in the full order space. We still observe that OpInf also outperforms the Markov neural operators (MNO) as proposed in \cite{likovachki} in terms of accuracy and stability. MNO keeps up with the exact trajectory for the KS system until $t=50s$ while OpInf exceeds $t=70s$.

\begin{figure}[!h]
\centering
\includegraphics[scale=0.5]{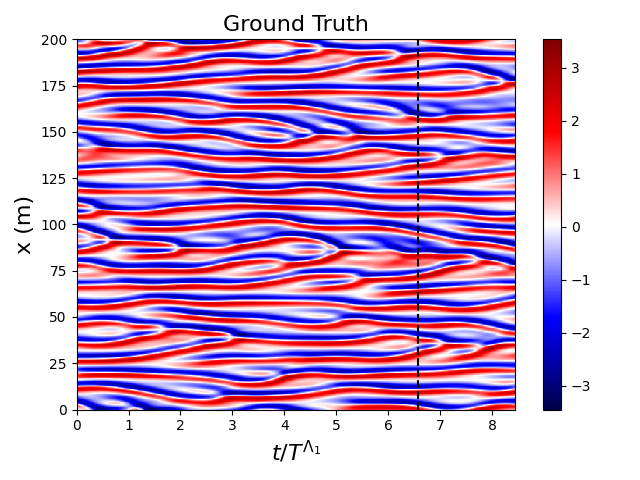} \hspace{-0.3cm}
\includegraphics[scale=0.5]{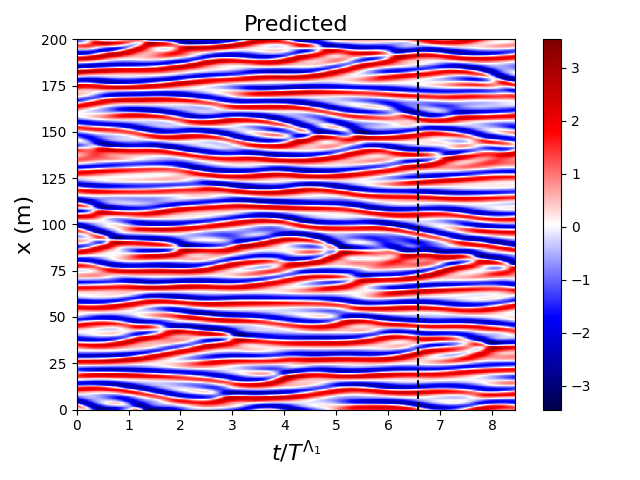}
\caption{Contour plots of the spatiotemporal forecast of the Kuramoto-Sivashinsky system with $160$ reduced dimensions. The dashed line indicates the visual limit for which OpInf prediction and ground truth match. Color bars show absolute values of the discretized system state variable $u(x,t)$.}
\label{KS_approx_vs_exact}
\end{figure}

Fig. \ref{KS_approx_vs_exact} compares the contour plots of one of the best OpInf approximate solutions (considering all the solutions corresponding to each initial condition) with the ground truth solution of the KS chaotic system. The OpInf ROM can accurately predict the system dynamics for nearly $6.7$ Lyapunov time units (cf. black dashed lines) which considerably outstrip the results achieved in \cite{vlachas2020backpropagation}. 

\begin{figure}[!h]
\centering
\begin{tabular}{cc}
\includegraphics[scale=0.5]{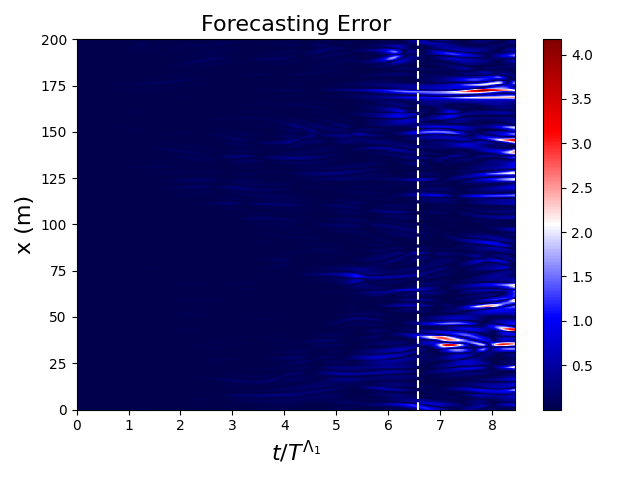} & \hspace{-0.7cm}
\includegraphics[scale=0.5]{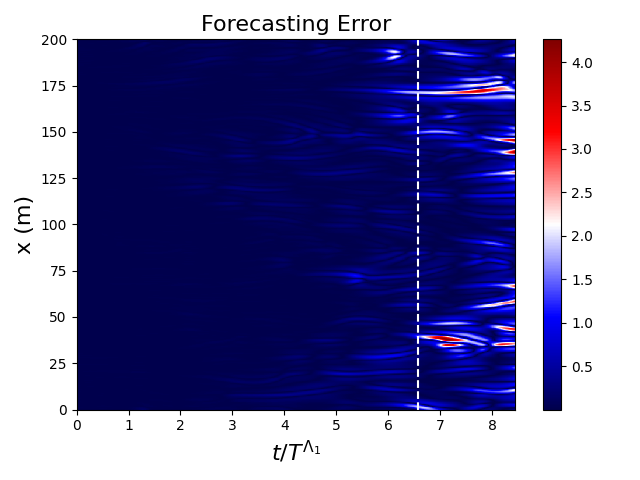} \\
a) Pointwise error & \hspace{-0.7cm} b) NRSE
\end{tabular}
\caption{Contour plots of the error measure between the ground truth and the approximate solution of the Kuramoto-Sivashinsky system with $160$ reduced dimensions: a) pointwise error; b) NRSE. The color bar indicates absolute values (e.g., $1.0$ represents $100\%$ deviation). The dashed line indicates the visual limit for which OpInf prediction and ground truth match.}
\label{KS_error}
\end{figure}

Fig. \ref{KS_error} shows the spatiotemporal contour plots of the approximation errors of the OpInf ROM based on two distinct metrics: a) pointwise error and b) NRSE. The white dashed lines point out how far the reduced model can keep errors at an extremely low level. Such noticeable performance highlights how physics-inspired ML architectures can outperform deep learning state-of-the-art models such as backpropagation and reservoir computing RNN architectures in extrapolation regime.

\begin{figure}[!h]
\centering
\includegraphics[width=8.0cm,height=3.7cm]{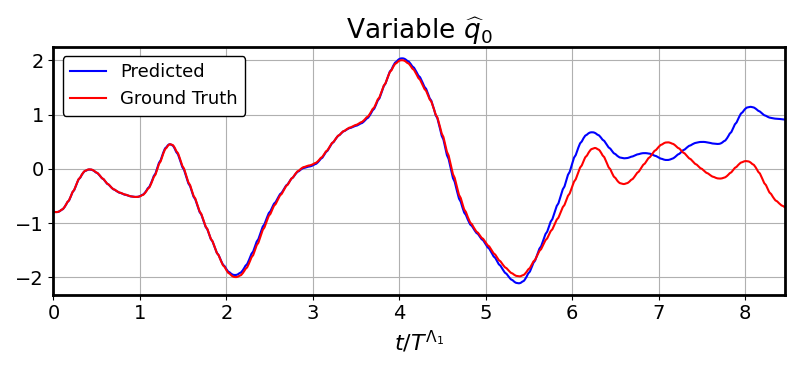}
\includegraphics[width=8.0cm,height=3.7cm]{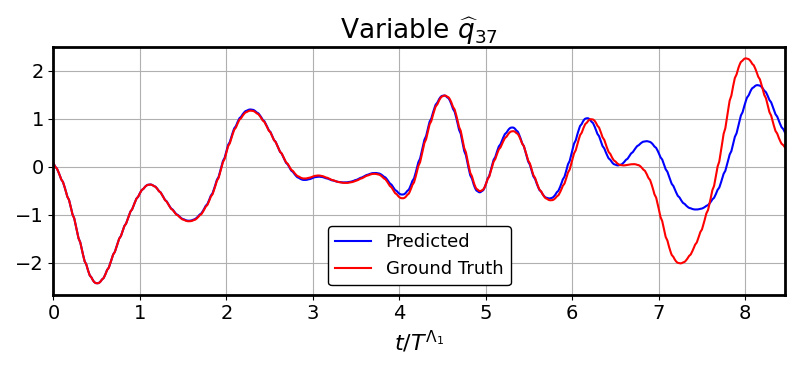}
\includegraphics[width=8.0cm,height=3.7cm]{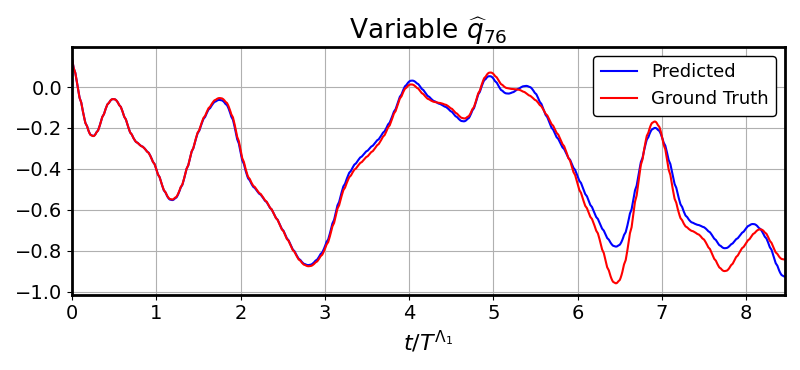}
\includegraphics[width=8.0cm,height=3.7cm]{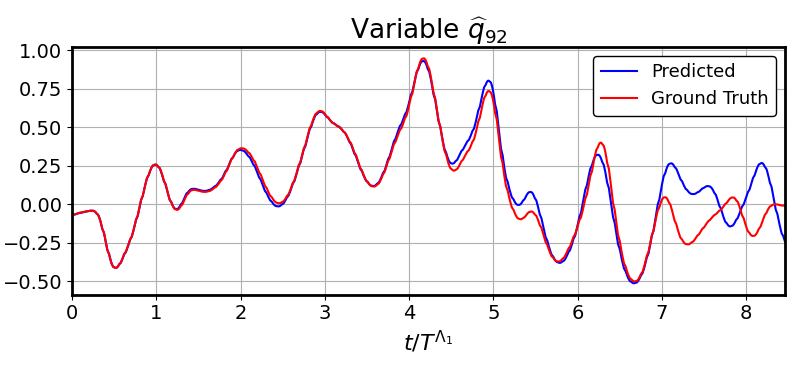}
\includegraphics[width=8.0cm,height=3.7cm]{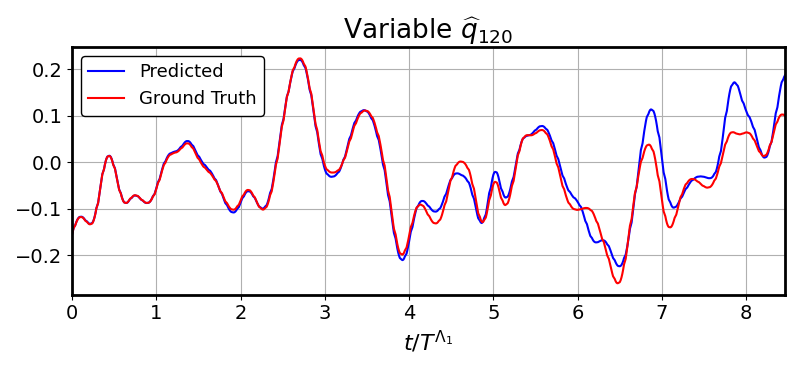}
\includegraphics[width=8.0cm,height=3.7cm]{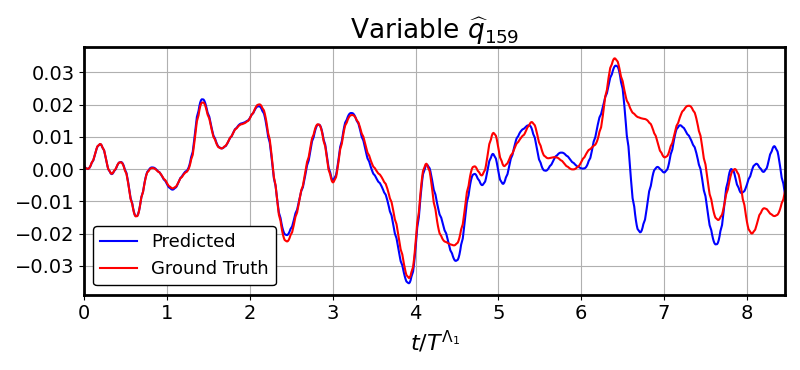}
\caption{Six temporal eigenmodes out of the total $160$ modes for the Kuramoto-Sivashinsky system: predicted (blue) vs. ground truth (red) variables.}
\label{ks_modes}
\end{figure}

Fig. \ref{ks_modes} shows the comparison between approximate and reference temporal eigenmodes for six distinct integrated time-series in the KS system latent space. The plots evidence the capacity of OpInf ROM to match the reference eigenmodes during long-term forecasting even for the lowest energetic modes (e.g., eigenvectors 120 and 159).

We notice in Fig. \ref{ks_modes} that the latent space representation of the KS data produces an increasing oscillatory pattern as the eigenmodes become less and less energetic. In contrast, the Lorenz 96 system with $F=10$, for instance, displays dominant high frequency oscillations in every temporal eigenmode since it is represented in the full order space (see Fig. \ref{lorenz_96_modes_F=8,10}b).

From the computational efficiency point of view, a single realization of the OpInf ROM takes $278\,s$ to prompt an output for the Kuramoto-Sivashinsky system, including the evaluation of the $\D^T\,\D$ matrix. We should keep in mind that for the full cycle of grid search optimization, the computations do not include the $\D^T\,\D$ assembling since this matrix is computed once and for all realizations. The authors in \cite{vlachas2020backpropagation} report times higher than $1,200\,s$ to perform the same task using RC-ESN with $3,000$ nodes. We should also stress that the complete hyper optimization workflow in this work with no extra tuning takes approximately $2.36\,h$ to run, including the pre and post-processing operations. All the numerical tests we performed were executed in a 20-core Intel Xeon CPU.

\begin{figure}[!h]
\centering
\begin{tabular}{ccc}
\hspace{-1.1cm}
\includegraphics[scale=0.42]{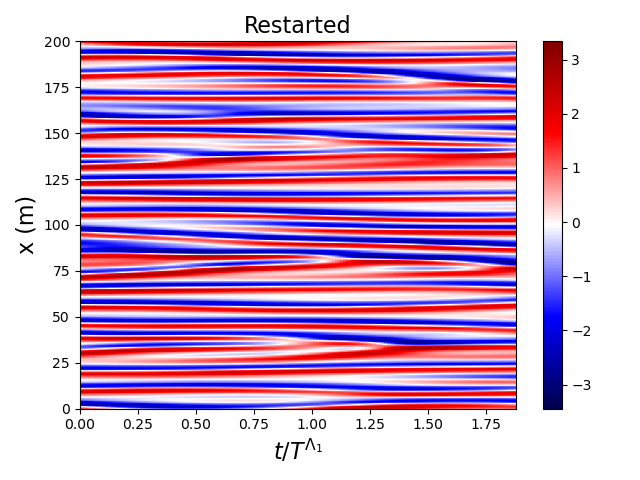} & \hspace{-1.1cm}
\includegraphics[scale=0.42]{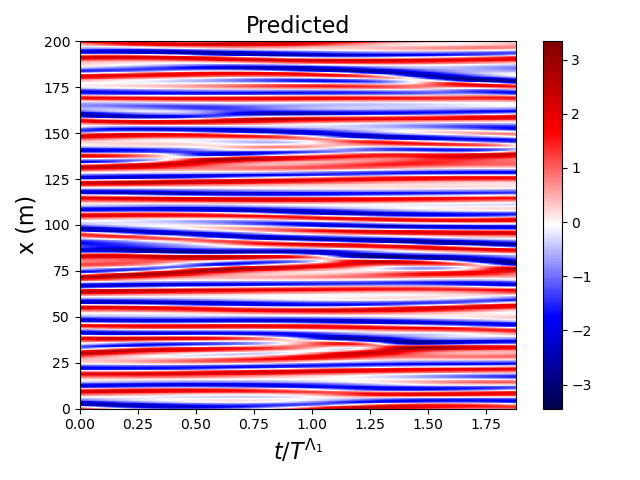} & \hspace{-1.1cm}
\includegraphics[scale=0.42]{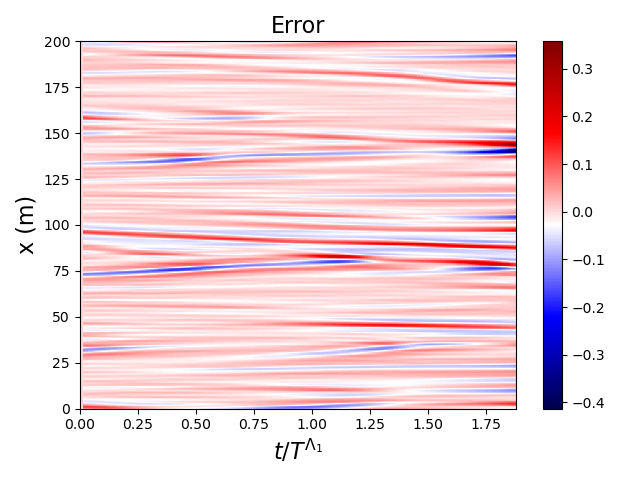} \\
\hspace{-1.1cm} a) & \hspace{-1.1cm} b) & \hspace{-1.1cm} c) 
\end{tabular}
\caption{Contour plots of the solution $u(x,t)$ of the KS system: a) restarted; b) OpInf (predicted); c) absolute pointwise error. Color bars show absolute values of the discretized system state variable $u(x,t)$.}
\label{KS_restarted_vs_predicted}
\end{figure}

As a final assessment, we check the physical consistency of the OpInf ROM. We select a point in the state-space of the ROM where the predicted solution and the ground truth already diverged ($\approx$ $6.8$ Lyapunov times in Fig. \ref{KS_error}) and use it as an initial condition for the discretized KS equation (i.e., the reference system). We then integrate the discretized equation in time for $20\,s$ ($\approx 1.88$ Lyapunov times) and compare the output with the OpInf solution. We call the time-integrated solution based on the reference system the restarted solution. Fig. \ref{KS_restarted_vs_predicted} shows the spatiotemporal representation of the restarted and OpInf solutions together with the absolute pointwise error between both solutions. We observe an excellent agreement between both KS systems' answers, with errors kept very small on most of the $(x,t)$ domain except for a few dark blue and red stripes.

\begin{figure}[!h]
\centering
\includegraphics[width=8.0cm,height=3.68cm]{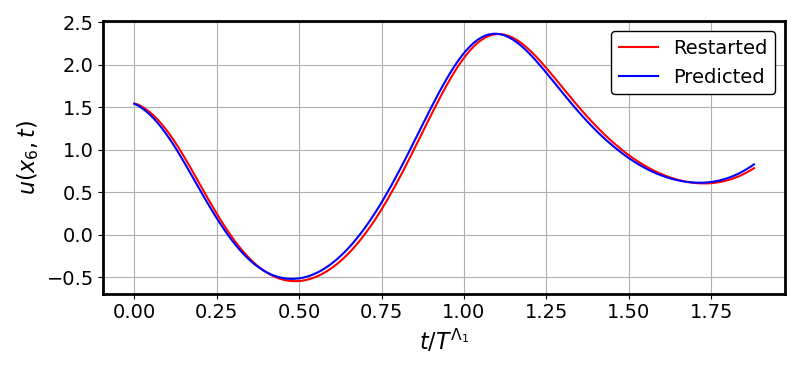}
\includegraphics[width=8.0cm,height=3.68cm]{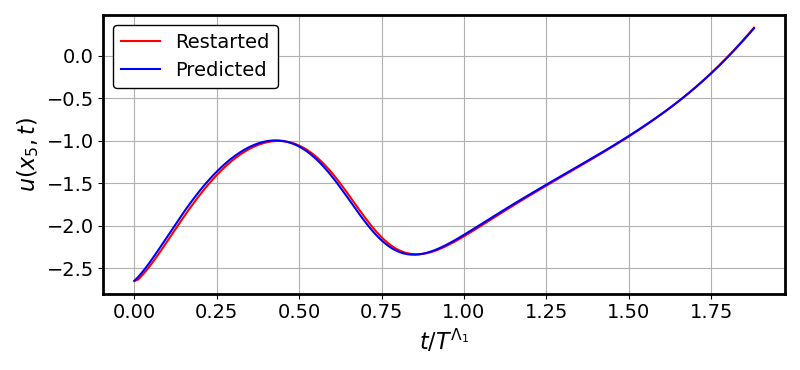}
\includegraphics[width=8.0cm,height=3.68cm]{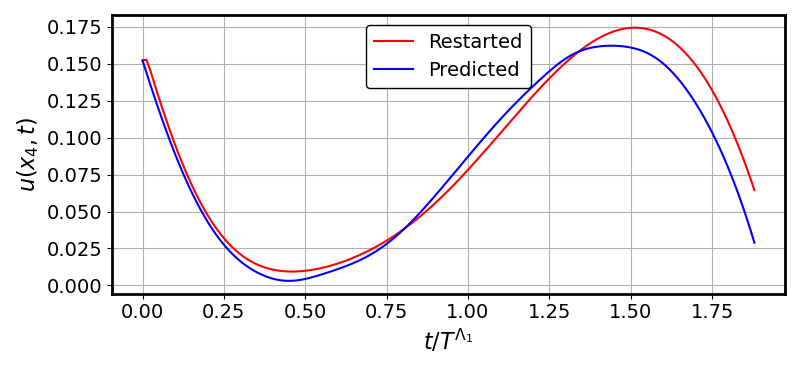}
\includegraphics[width=8.0cm,height=3.68cm]{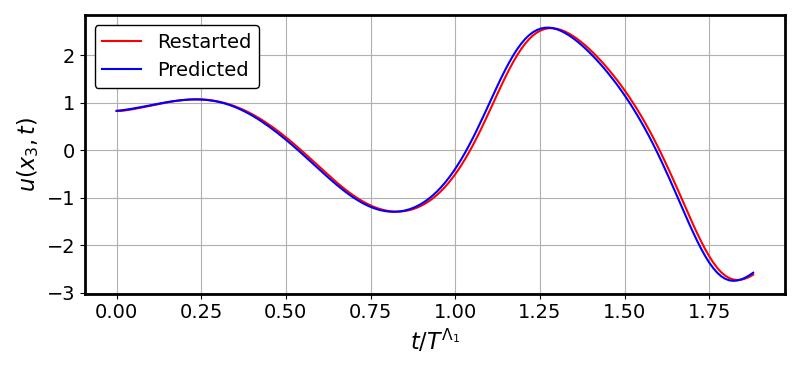}
\includegraphics[width=8.0cm,height=3.68cm]{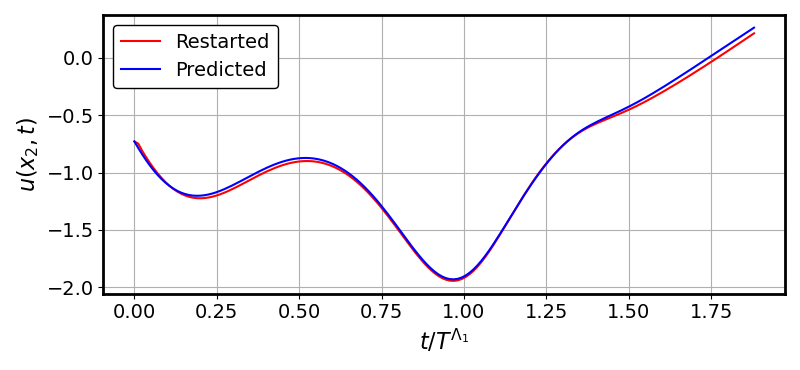}
\includegraphics[width=8.0cm,height=3.68cm]{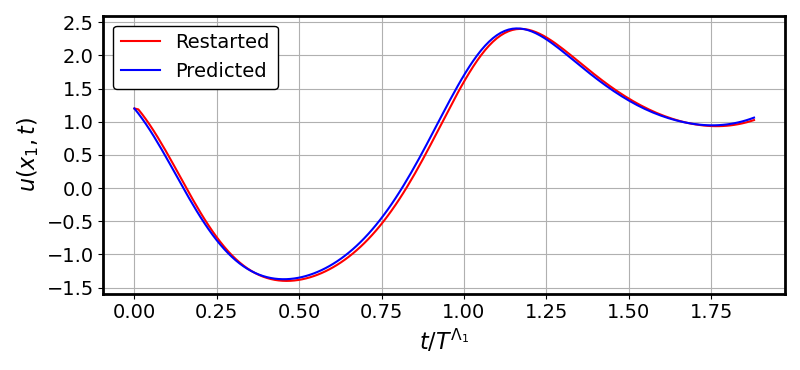}
\caption{Restarted vs. OpInf ROM solutions of the KS system at regularly spaced grid points: $u(x_{1},t),\dots,u(x_{6},t)$. The top and bottom plots correspond to the solutions at the boundaries of the KS domain $u(x_{6} = L,t)$ and $u(x_{1} = 0,t)$, respectively.}
\label{tseries_restarted_vs_opinf}
\end{figure}

Fig. \ref{tseries_restarted_vs_opinf} compares the restarted and the OpInf ROM solutions at six regularly spaced grid points. These plots correspond to horizontal slices (time-series) of the contour plots in Fig. \ref{KS_restarted_vs_predicted}. Comparing the restarted and the OpInf solutions on these plots, we confirm a close similarity between them, especially at the boundaries of the KS domain $u(0,t)$ and $u(L,t)$ (cf. first and last plots from top to bottom in Fig. \ref{tseries_restarted_vs_opinf}). This result indicates that OpInf is physically meaningful since its outputs are consistent with the state-space generated by the reference system including its boundary conditions.

\section{Conclusion}\label{sec:conclusions}
In this paper we develop surrogate models based on the Non-intrusive Reduced Order Modeling via Operator Inference (OpInf) method to simulate benchmark chaotic systems, namely, Lorenz 96 and Kuramoto-Sivashinsky (KS) equation. We also propose a parallel scheme for decomposing the matrix product $\D^T\,\D$ on the left hand side of Equation \eqref{eq:8} using a batch-wise approach as detailed in Appendix \ref{parallel_eval}. The OpInf method operates on data generated by high-fidelity simulations organized in a sequence of snapshots. PCA decomposition reduces the dimensionality of the input dataset and projects the original snapshots on the latent space as a system of ODEs. OpInf postulates a quadratic polynomial structure for the reduced approximation space and performs a ridge regression scheme to find the reduced operators that best fit the dynamics in the latent space. Regularization is added to the optimization process to avoid overfitting.

The OpInf shows remarkable forecasting capabilities for both dynamical systems, with two setup configurations for Lorenz 96 and one for the KS system. We observe robust time-series predictions in latent space outperforming state-of-the-art ML techniques such as ESN-RC, Unitary cell, LSTM, and GRU analyzed in \cite{vlachas2020backpropagation} by $43\%$. The OpInf performance also excels in the Markov Neural Operator (MNO) method \cite{likovachki} based on Fourier Neural Operators (FNO) being $49\%$ more accurate in representing the KS system trajectory. It is worth noticing that for Lorenz 96 system with 40 degrees of freedom, OpInf reached VPT at $19.15$ and $15.39$ Lyapunov time units in the best case for forcing terms $F=8$ and $F=10$, respectively, while the best forecasting model in \cite{vlachas2020backpropagation} topped at $2.31$ and $2.35$ Lyapunov time units. Regarding the KS system, OpInf performs remarkably well compared with the results in \cite{vlachas2020backpropagation} and \cite{likovachki}, matching ground truth simulations for as long as \textcolor{blue}{$7.01$} Lyapunov time units in the best case using $160$ degrees of freedom. Referring to Fig. 12a in \cite{vlachas2020backpropagation}, we notice that their best surrogate model defined in full space dimension ($512$ degrees of freedom) reached $\approx 4.9$ Lyapunov time units as VPT. Numerical experiments for the KS system also revealed that OpInf ROM produces physically meaningful solutions consistent with the state-space generated by the discretized KS equation, as depicted in Figs. \ref{KS_restarted_vs_predicted} and \ref{tseries_restarted_vs_opinf}. In this case, the OpInf dynamics show close correspondence with the solution generated by the reference system using an OpInf unpaired physical state as the initial condition.

Although direct clock measuring is not the best criterion for creating a performance index analysis for comparing algorithms running on different hardware, the computational efficiency assessment reported at the end of sections \ref{res:lor} and \ref{res:ks} gives an insight into how OpInf can be much faster than its black-box ML competitors to solve the problems we considered in this work for a single algorithm realization. For the Lorenz 96 case, OpInf performs a single realization in a matter of $3.8\,s$, while it takes $278\,s$ in the case of the KS system. LSTM, RC-ESN, and other similar algorithms reported in \cite{vlachas2020backpropagation} demand much more computational power and time to perform the same task. The OpInf ROM proves to be an extremely parsimonious algorithm that delivers quite accurate results.

We corroborate the many attractive features of OpInf. It is non-intrusive, generalizable (in the sense that it can approximate a large class of PDE operators with linear-quadratic structure), explainable, physics-informed, and computationally efficient. Moreover, it is based on a sound mathematical background (Koopman’s theory), does not depend on a gradient descent algorithm, uses standard linear algebra tools, and needs only a few regularizing parameters. The regularization strategy in the least-squares problem is a key ingredient for the OpInf success, but it is quite tricky. The Tikhonov-based regularization method used in OpInf is obtained through a grid search over the space of the regularization parameters, which makes the process of finding the global minimum pretty challenging. Keeping time integration errors bounded is another crucial point we address using SciPy's LSODA algorithm. We should also remark that the OpInf capability of producing good approximation results strongly relies on the availability of large training datasets generated by numerical solvers. Such requirement is essential for properly computing time derivatives and then fitting the model. We expect that for challenging real-world applications such as turbulent flows, the storage demand will be even higher. In such cases, we consider employing highly accurate differentiation methods, as proposed in \cite{Lele1992}, or even representation models \cite{osti}, which are neural networks trained to represent datasets as a function of spatiotemporal coordinates and harness their automatic differentiation feature to compute time-derivatives straightforwardly.

As we are interested in tackling turbulent flows in the future, we plan to find ways to build closure mechanisms for the OpInf ROM in latent space aiming at capturing the energy of hidden scales smeared by the PCA decomposition. Closure mechanisms could be straightforwardly plugged into the OpInf least-squares problem through the $\mathbf{\widehat{B}}$ forcing operator (cf. Equation \eqref{eq:5}) producing a ROM closure. We also want to address the limitations of the OpInf grid search regularization technique by exploring convex and nonconvex optimization approaches that best fit the regularization parameters by means of approximations and exploration of equivalent formulations.

\section*{Acknowledgment}
ACNJ acknowledges the Brazil's IBM Research laboratory for supporting this work.

\bibliographystyle{unsrt}  
\bibliography{references}

\begin{appendices}

\section{Pseudocode for the parallel evaluation of $\D^T\D$}
\label{parallel_eval}

\begin{empheq}[box=\fbox]{align*}
    	&\sum^{N_b}_{j=0} n_j = N\\
    	&\D^T\,\D = \Ma = \sum^{N_b}_{j=0} \D^{T}_j\D_j,\\
    	&N_a = 0\\
    	&\Ma = \mathbf{0}\\\\
    	&\rhd\textit{Batch-wise loop}\\
    	&\textbf{for} \,\, j \,\, \textbf{in} \,\,0:N_b\,\,\textbf{do}\\\\
    	&\qquad\rhd\textit{Dispatch it into a MPI process}\\
    	&\qquad \Q_j = \Q[N_a:N_a + n_j],\\
    	&\qquad \D_j = [\mathbf{1}\,\,\, \Q_j\,\,\,\Q_j \otimes \Q_j]\\\\
    	&\qquad\rhd\textit{Accounting the batch in the global matrix}\\
    	&\qquad \Ma \xleftarrow[]{}  \D_{j}^T \D_j\\
        &\qquad N_a \xleftarrow[]{} n_j\\
\end{empheq}\\
where $N$ is the total number of time samples, $N_b$ the total number of batches and $\Q$ is the snapshots matrix obtained via PCA.

\section{$\Gamma$ function used for constructing the regularization matrix $\mathbf{\Gamma}$}
\label{lambda_gamma}

\begin{equation*}
    \Gamma:\left\{ 
    \begin{array}{ll}
    &\textit{diag}(\mathbf{\Gamma})_i = \lambda_1, \,\,\text{if}\,\,i = 0 \\
    &\textit{diag}(\mathbf{\Gamma})_i = \lambda_2,\,\,\text{for}\,\,i\,\,\in\,\,[1, r] \\ 
    &\textit{diag}(\mathbf{\Gamma})_i = \lambda_3,\,\,\text{for}\,\,i\,\,\in\,\,[r, r + r(r+1)/2] \\
    &\textit{diag}(\mathbf{\Gamma})_i = \lambda_4,\,\,\text{for}\,\,i > r(r+1)/2
    \end{array}
    \right.
\end{equation*}

\end{appendices}

\end{document}